\documentclass[11pt]{article}
\usepackage[T1]{fontenc}
\usepackage{CJKutf8}

\newcommand{\mandarin}[1]{\begin{CJK}{UTF8}{gbsn}#1\end{CJK}}
\newcommand{\korean}[1]{\begin{CJK}{UTF8}{mj}#1\end{CJK}}

\usepackage{bm}
\usepackage{geometry}
\usepackage{fullpage}
\usepackage{url}
\usepackage{gb4e}
\noautomath
\usepackage[english]{babel}
\usepackage{graphicx}
\usepackage{color}
\usepackage{rotating}
\usepackage{tabularx}
\usepackage{natbib}
\usepackage{amsmath}
\usepackage{tipa}
\usepackage{comment}
\includecomment{comment}
\usepackage{pgfplots}

\usetikzlibrary{arrows,shapes,shadows,decorations.pathmorphing,backgrounds,positioning,fit,petri, automata}
\usepackage[absolute,overlay]{textpos}

\usepackage[unicode=true,pdfusetitle,
 bookmarks=true,bookmarksnumbered=true,bookmarksopen=true,bookmarksopenlevel=2,
 breaklinks=true,pdfborder={0 0 1},backref=false,colorlinks=true,citecolor=blue,urlcolor=blue,linkcolor=blue]
 {hyperref}

\usepackage{subcaption}

\usepackage{tikz}
\usepackage{tikz-3dplot}

\usepackage[theme=default, charsperline=90]{jlcode}

\title{Vector Space Morphology with Linear Discriminative Learning}
\author{Yu-Ying Chuang, Mihi Kang, Xuefeng Luo, and R. Harald Baayen \\ University of T\"{u}bingen} 
\date{July, 2021}

\begin{document}
\maketitle

\begin{center}
    {\bf Abstract}
\end{center}

\noindent
This paper presents three case studies of modeling aspects of lexical processing with Linear Discriminative Learning (LDL),  the computational engine of the Discriminative Lexicon model \citep{Baayen:Chuang:Shafei:Blevins:2019}. With numeric representations of word forms and meanings, LDL learns to map one vector space onto the other, without being informed about any morphological structure or inflectional classes. The modeling results demonstrated that LDL not only performs well for understanding and producing morphologically complex words, but also generates quantitative measures that are predictive for human behavioral data. LDL models are straightforward to implement with the JudiLing package \citep{Luo:Chuang:Baayen:2021}. Worked examples are provided for three modeling challenges: producing and understanding Korean verb inflection, predicting primed Dutch lexical decision latencies, and predicting the acoustic duration of Mandarin words.

\vspace*{0.5cm}

\vspace*{0.5cm}

\section{Introduction}

Linear Discriminative Learning (LDL) is the statistical engine implementing the mappings between form and meaning in a computational model of the mental lexicon, the `Discriminative Lexicon' (DL) laid out in \citet{Baayen:Chuang:Shafei:Blevins:2019}. Given numeric representations for words' forms as well as numeric representations for their meanings, the mathematics of multivariate multiple regression are used to transform form vectors into meaning vectors (comprehension), and to transform meaning vectors into form vectors (production).  The beta coefficients of the regression models are equivalent to the weights on the connections in fully connected simple networks with input units, output units, and no hidden layers, when these network is trained to perfection \citep{Chuang:Bell:Banke:Baayen:2020,ShafaeiBajestan:2021}.  In other words, the regression coefficients represent the best estimates (in the least squares sense) of the endstate of incremental learning.  In this study, we focus on this endstate.

Although extremely simple, we have found the DL and its implementations of vector space morphology with LDL surprisingly effective for modeling a wide range of phenomena in lexical processing \citep[see][for an overview]{Chuang:Baayen:2021}. In what follows, we present three new case studies illustrating the potential of vector space morphology for morphological theory and the study of lexical processing.  We first illustrate how LDL can be used to model Korean verb inflection.  Our goal here is to show that computationally, the algorithms of LDL work for a non-trivial dataset of a language with complex morphology.  We then show how LDL can be used to predict morphological priming in auditory comprehension, using data from Dutch \citep[Experiment~1 of ][]{Creemers:2019}.  Here our goal is to demonstrate that our approach offers predictive precision at the item level for data that prima facie would seem to support theories positing obligatory morphological decomposition as the first step in the lexical processing of complex words.  Our third case study addresses the production of Mandarin words, focusing on acoustic duration in spontaneous speech as response variable.  The goal of this final study is to illustrate how measures can be derived from LDL networks that make it possible to enhance prediction accuracy beyond what is possible with standard predictors such as word length in phones and frequency of occurrence.  The general discussion provides reflection on the implications of our approach for morphological theory, theories of lexical processing, and experimental design of studies on morphological processing.

\section{Vector space morphology for Korean verbs}\label{sec:korean}

Korean is generally considered to be a language isolate, possibly distantly related to Altaic languages such as Japanese.  It has a complex morphology that challenges linguistic analysis in many ways, even though it is generally described to be agglutinative.  The Korean verb system is especially rich.  Although there is no agreement about how to best analyse Korean verbs \citep[see][for detailed discussion]{yang1994morphosyntactic}, the seven categories \citep[roughly following][]{sohn2013korean} listed in Table~\ref{tab:overview} provide an overview of the factors that jointly determine a verb's form.
  
\begin{table}[h]
\caption{Inflectional and derivational dimensions of the Korean verb system. The values in italics pertain to example (1).} 
\vspace*{0.2\baselineskip}
\centering
\begin{tabular}{lll} \hline
    derivation          & (1) & active (unmarked), causative, {\em passive}    \\
    subject honorifics  & (2) & plain (unmarked), {\em honorific}       \\
    tense               & (3) & present (unmarked), {\em past}, future  \\
    modality            & (4) & neutral (unmarked), volition, {\em conjecture}  \\ 
    addressee honorifics& (5) & plain (unmarked), {\em formal}, polite, intimate \\
    mode                & (6) & neutral (unmarked), subjunctive, {\em retrospective},  \ldots \\
    illocutionary force & (7) & declarative, {\em inquisitive}, imperative, propositive \\ \hline
\end{tabular}
\label{tab:overview}
\end{table}

Subject honorifics express the relative status of the speaker and the subject of the clause.  If the subject of the sentence has superior status, the subject typically is an older relative, a stranger, an employer, or a teacher.  Addressee honorifics honor the relative status of speaker and interlocutor, they are used to show respect towards a speaker's or writer's audience.  Addressee honorifics are also referred to as speech levels, and we will use this terminology. Up to six speech levels have been distinguished in the literature but in Table 1 we only list the four most common ones. The category labeled `mode' comprises a great many nuances that are realized by means of a wide range of exponents.  The language has a four-way contrast for different kinds of illocutionary force. Example~(1) presents a sentence with a verb on which all these seven inflectional categories are realized, the numbers refer to the inflectional and derivational dimensions listed in Table~\ref{tab:overview}, and the values for these dimensions are presented in italics in this table.

\begin{exe}
\ex 
\korean{그 분이 잡히시었었겠습디까}? 
\vspace*{-0.6\baselineskip}
\gll ku	pwun-i		cap-hi-si-ess-ess-keyss-sup-ti-kka? \\
 the	person-NM	catch-(1)-(2)-(3-3)-(4)-(5)-(6)-(7)\\\vspace*{-0.6\baselineskip}
\trans `Did you feel (guess) that he had been caught?' \citep{sohn2013korean}
\end{exe}

\noindent
Most Korean verb forms do not overtly express all seven morphological categories.  When particular categories are not expressed, the ordering of the inflectional exponents for the other categories remains the same as in (1). Table \ref{tab:example_forms} shows some possible combinations of the inflectional verb forms varying in speech levels (the first four examples) and illocutionary force (the other four examples at the bottom). 

\begin{table}[]
\caption{Examples of inflected variants of the verb \korean{만나}(manna), `to meet'. The top four examples vary in speech levels, the bottom four examples illustrate inflection for illocutionary force. Dots represent approximate exponent boundaries. SH: subject honorifics; SL: speech level.}
\label{tab:example_forms}
\vspace*{0.2\baselineskip}
\centering
\begin{tabular}{lll} \hline
manna-si.ess.up.nita    & \korean{만나셨습니다} &	stem-honorific(SH)-past-formal(SL)-declarative \\
manna-si.ess.eyo		& \korean{만나셨어요}   &   stem-honorific(SH)-past-polite(SL)-declarative \\
manna-si.ess.e		    & \korean{만나셨어}     &	stem-honorific(SH)-past-intimate(SL)-declarative \\
manna-si.ess.ta		    & \korean{만나셨다}     &	stem-honorific(SH)-past-plain(SL)-declarative \\ \hline
manna-nta		        & \korean{만난다}       & 	stem-plain(SH)-present-plain(SL)-declarative \\
manna-ni    		    & \korean{만나니}       & 	stem-plain(SH)-present-plain(SL)-inquisitive \\
manna-la		        & \korean{만나라}       & 	stem-plain(SH)-present-plain(SL)-imperative \\ 
manna-ca		        & \korean{만나자}       &	stem-plain(SH)-present-plain(SL)-propositive \\ \hline
\end{tabular}
\end{table}

As mentioned above, theories vary in terms of how to decompose Korean verbs into stems and exponents \citep[e.g.,][]{lee1989korean,choi1929uli,martin1960korean,yang1994morphosyntactic}. The segmentation in (1) follows \citet{sohn2013korean}. The segmentation of the inflectional features and their semantic interpretation differ from scholar to scholar. For example, a verbal suffix sequence such as \korean{-(ㅅ)읍니다}(-(s)upnita) is analyzed as: 
\begin{enumerate}
    \item sup-ni-ta  \korean{-습니다}, formal speech level+indicative mood+declarative \citep{sohn2013korean};
    \item upni-ta    \korean{-읍니다}, formal speech level+declarative \citep{yang1994morphosyntactic};
    \item up-nita    \korean{-읍니다}, formal speech level+declarative (\citet{choi1929uli}, cited in \citet{yang1994morphosyntactic});
    \item up-ni-ta   \korean{-읍니다}, formal speech level+indicative mood+declarative \citep{lee1989korean}.
\end{enumerate}
Although Korean is described as agglutinative, i.e., with a morphological system that uses sequences of affixal morphemes as building blocks, the above examples clarify that a strict morpheme-based analysis is not straightforward.   Some exponents are clearly fusional, such as \korean{-라}(-la) in \korean{만나라} (manna-la), which realizes both plain speech level and imperative illocutionary force.  Other exponents that realize both plain speech level and illocutionary force are \korean{-니}(-ni), \korean{-자}(-ca), and \korean{-ㄴ다} (-nta).  Both stems and exponents may show allomorphy or even suppletion.  Thus, the infinitive of the verb `to eat' has two allomorphs, \korean{먹}(mek), and \korean{먹어}(meke). Likewise, \korean{-ㄴ다}(-nta, plain speech level, declarative illocutionary force) is found after verb stems ending in vowels, but it appears as \korean{-는다}(-nunta) after consonants. Stem suppletion is found for, e.g., the verb `to eat', which when used with plain honorific speech requires the allomorph \korean{먹}(mek), but when used with subject honorifics is realized as \korean{드시}(tusi). 
Unsurprisingly, in second language teaching of Korean, learners are guided to pick up a range of low-level markers for inflectional functions, without calling explicit attention to theoretical position classes as exemplified in (1).  
 
The non-trivial morphology of Korean raises the question of whether the simple linear mappings of discriminative learning can actually master this system, without ever having to define morphemes or exponents.  To address this question, the present study focuses on verb forms realizing combinations of the four most frequently used inflectional categories: subject honorifics (labelled as `honorifics' in our dataset), tense, speech level, and illocutionary force.  The feature values that we implemented are those listed above in Table~\ref{tab:overview}.   Our dataset also includes some derived words realizing passives or causatives, as these features are commonly used in daily speech.  The combinations of these features result in 59 inflected forms for a standard verb.   Semantically impossible combinations are not included (e.g., past propositive or future imperative).  In this dataset, which in total comprises 27,258 word forms of in all 462 verbs, words' forms are represented by syllables.  These syllables do not necessarily coincide with Hangul characters.  Although Hangul characters are designed to represent syllables, due to assimilation, contraction, and resyllabification, the spoken forms of words can diverge considerably from the Hangul orthography. We opted for phonological transcriptions that follow the spoken forms.   Table~\ref{tab:example} presents four rows of our dataset, which provides the Hangul written form, a phonemic representation with underscores indicating syllable boundaries, an identifier for the verb lexeme, and the feature specifications for honorifics, tense, speech level, and illocutionary force.  In our dataset, differences in intonation, which can distinguish between words that are otherwise homophones but that have different illocutionary force (e.g., \korean{먹어요} (mekeyo), `to eat', present tense, polite speech level, declarative/inquisitive/propositive/imperative), are not yet indicated.  

\begin{table}
\caption{Four Korean verb forms and their phonological and morphological properties that form the input for modeling with LDL.}
\label{tab:example}
\begin{small}
\begin{CJK}{UTF8}{mj}
\begin{tabular}{r|lllllll} \hline
  & Hangul      & Word              & Lexeme & Honorifics & Tense   & SpeechLevel & IllocutionaryForce\\	\hline
1 & 고릅니다    & go\_rUm\_ni\_da   & gorUda & plain      & present & for         & dec \\
2 & 고릅니까    & go\_rUm\_ni\_kka  & gorUda & plain      & present & for         & inq \\
3 & 고르십시오  & go\_rU\_sip-\_syo & gorUda & hon        & present & for         & imp \\
4 & 고릅시다    & go\_rUp-\_si\_da  & gorUda & plain      & present & for         & pro \\ \hline
\end{tabular}
\end{CJK}
\end{small}
\end{table} 

In what follows, we will introduce the main concepts behind LDL step by step, without going into mathematical detail, but providing the code for setting up and running the model using the {\bf JudiLing} package for the {\tt Julia} language \citep{Luo:Chuang:Baayen:2021}.\footnote{For installation of the package and the packages that it depends on, see the on-line manual at \url{https://megamindhenry.github.io/JudiLing.jl/stable/\#Installation}.} The choice for Julia as programming language is motivated by the fact that this language is extremely well optimized for numerical calculation.  As a consequence, building and evaluation with {\bf JudiLing} is much faster than is possible with the {\bf WpmWithLdl} package for {\tt R}.\footnote{The latest version of this package ({\tt WpmWithLdl\_1.4.6.tar.gz}) is available at \url{https://osf.io/xq92s}.}.  

Before starting the actual modeling, after having started up Julia, we first load the {\bf JudiLing} package and the packages for handling csv files and data frames.  The following code snippet assumes that our dataset, {\tt korean.csv}, is available in the current working directory.
\\ \ \\
\begin{jllisting}
using JudiLing, CSV, DataFrames
korean = DataFrame(CSV.File("korean.csv"));
\end{jllisting}
\noindent
In order to set up mappings between words' forms and their meanings, we need to define numeric representations for form and meaning.  We first consider setting up a matrix with row vectors that represent words' forms.  There are many ways in which form vectors can be defined.  In the present study, we set up form vectors that specify which pairs of consecutive syllables occur in a word.  That is, we build a form matrix $\bm{C}$ with as many rows as there are words, and with as many columns as there are syllable pairs.  We indicate the presence of a syllable pair with 1, and its absence with 0. Constructing this matrix with {\bf JudiLing}'s {\tt make\_cue\_matrix} function is straightforward. 
\begin{jllisting}
C_obj = JudiLing.make_cue_matrix(korean, grams=2, target_col=:Word, tokenized=true,
                                 sep_token="_", keep_sep=true);
\end{jllisting}
The directive {\tt grams=2} tells the model to construct di-syllables, the directive {\tt target\_col} specifies in which column of the dataset word form information can be found, and the {\tt sep\_token} directive specifies that the underscore represents the syllable boundary marker. The object returned by {\tt make\_cue\_matrix}, {\tt C\_obj}, contains several useful data structures, of which the $\bm{C}$ matrix ({\tt C\_obj.C}) is of interest to us here.   
\begin{jllisting}
JudiLing.display_matrix(korean, :Word, C_obj, C_obj.C, :C, nrow=4, ncol=9)
\end{jllisting}

\begin{flushleft}
\begin{footnotesize}
\begin{tabular}{r|lccccccccc} \hline
	& & \#\_go & go\_rUm & rUm\_ni & ni\_da & da\_\# & ni\_kka & kka\_\# & go\_rU & rU\_sip-\\
	\hline
	1 & go\_rUm\_ni\_da & 1 & 1 & 1 & 1 & 1 & 0 & 0 & 0 & 0 \\
	2 & go\_rUm\_ni\_kka & 1 & 1 & 1 & 0 & 0 & 1 & 1 & 0 & 0 \\
	3 & go\_rU\_sip-\_syo & 1 & 0 & 0 & 0 & 0 & 0 & 0 & 1 & 1 \\
	4 & go\_rUp-\_si\_da & 1 & 0 & 0 & 0 & 1 & 0 & 0 & 0 & 0 \\ \hline
\end{tabular}
\end{footnotesize}
\end{flushleft}

The second modeling step is to construct the matrix $\bm{S}$ that represents words' meanings in numeric form, as in distributional semantics \citep{Landauer:Dumais:1997,mikolov2013distributed,Mitchell:Lapata:2008}.  These vectors can be extracted from corpora using standard methods from computational linguistics.  In this study, however, we simulate these vectors, but we make sure that they properly represent inflectional contrasts  \citep[see][for further examples]{Baayen:Chuang:Blevins:2018,Chuang:Loo:Blevins:Baayen:2019}. This is accomplished by summing the simulated semantic vectors of the verb stem and the simulated vectors representing the various inflectional features realized in a given inflected form.  Thus, the semantic vector of the first word {\tt go\_rUm\_ni\_da} is obtained as follows:
\begin{equation}
\overrightarrow{go\_rUm\_ni\_da} = \overrightarrow{gorUda} + \overrightarrow{plain} + \overrightarrow{present\vphantom{plain}} + \overrightarrow{for} + \overrightarrow{dec}.
\end{equation}

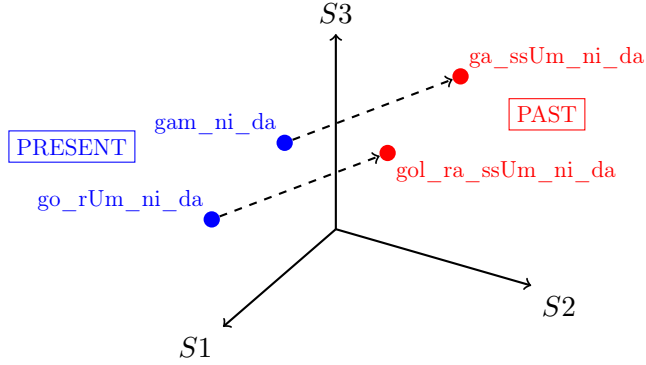
\begin{figure}
\centering
\tdplotsetmaincoords{60}{120}
\begin{tikzpicture}[scale=3,tdplot_main_coords]
  \coordinate (O) at (0,0,0);
  \draw[thick,->] (0,0,0) -- (1,0,0) node[anchor=north east]{$S1$};
  \draw[thick,->] (0,0,0) -- (0,1,0) node[anchor=north west]{$S2$};
  \draw[thick,->] (0,0,0) -- (0,0,1) node[anchor=south]{$S3$};

  \node[draw, scale=0.8, color=blue] () at (1.3,-0.6,0.9) {PRESENT};
  \node[draw, scale=0.8, color=red] () at (0.9,1.6,1.5) {PAST};

  \node[fill = blue, circle, scale=0.6] (b1) at (0.8,0.2,0.9) {};
  \node[fill = red, circle, scale=0.6] (s1) at (0.8,1.1,1.5) {};
  \draw[thick,->,dashed] (b1) node[anchor=south east,  color=blue, scale=0.8]{gam\_ni\_da} -- (s1) node[anchor=south west, color=red, scale=0.8]{ga\_ssUm\_ni\_da};

  \node[fill = blue, circle, scale=0.6] (b2) at (1.1,0,0.6) {};
  \node[fill = red, circle, scale=0.6] (s2) at (1.1,0.9,1.2) {};
  \draw[thick,->,dashed] (b2) node[anchor=south east, color=blue, scale=0.8]{go\_rUm\_ni\_da} -- (s2) node[anchor=north west, color=red, scale=0.8]{gol\_ra\_ssUm\_ni\_da};
\end{tikzpicture}

\caption{Vector addition shifts the position of a vector in  semantic space. In this example, the vectors of the two verbs in present tense forms (gam\_ni\_da `to go', go\_rUm\_ni\_da `to pick') are added with the vector of {\sc past}, which moves them from the {\sc present} area to the {\sc past} area.}
\label{fig:semantic_space}
\end{figure}

\noindent
All elementary simulated semantic vectors consist of random numbers following Gaussian distributions. By adding inflectional vectors to base vectors, the meanings of base words are shifted systematically, and in different ways that are specific to the inflectional meanings that are realized in conjunction with that base word.   Figure~\ref{fig:semantic_space} illustrates this concept for tense inflection.  By adding the vector of {\sc past} to the vectors of the verbs in the present tense, we move their positions from the area in space where present tense forms are located to the area in space where past-tense forms cluster. 

Simulating morphologically-aware semantic vectors with {\bf JudiLing} is straightforward. All we need to do is specify the dataset, the name of the column in the dataset that provides the verb stems (lexemes), and the names of the columns that specify inflectional feature values.  
\begin{jllisting}
S = JudiLing.make_S_matrix(korean, 
                           [:Lexeme], 
                           [:Honorifics, :Tense, :SpeechLevel, :IllocutionaryForce], 
                           ncol=size(C_obj.C, 2);
\end{jllisting}
The {\tt ncol} directive specifies the number of columns of $\bm{C}$, ensuring that the length of the form vectors and the semantic vectors are identical.  The upper-left corner of $\bm{S}$ is displayed as follows: 
\begin{jllisting}
JudiLing.display_matrix(korean, :Word, C_obj, S, :S, nrow=4, ncol=7)
\end{jllisting}

\begin{flushleft}
\begin{footnotesize}
\begin{tabular}{l|lrrrrrrr} \hline
	&  & S1 & S2 & S3 & S4 & S5 & S6 & S7\\
	\hline
	1 & go\_rUm\_ni\_da & -18.5863 & 6.61192 & 4.24111 & 2.01868 & 9.9817 & 14.7086 & 11.1827 \\
	2 & go\_rUm\_ni\_kka & -21.4642 & -2.87628 & 3.51315 & 4.98055 & 11.0686 & 15.6375 & 1.0228 \\
	3 & go\_rU\_sip-\_syo & -30.2516 & -2.66979 & 2.68477 & 1.4475 & 11.2798 & 6.68343 & 8.08859 \\
	4 & go\_rUp-\_si\_da & -23.2687 & -4.55982 & 0.965755 & 5.25853 & 12.1366 & 14.1624 & 1.43157 \\ \hline
\end{tabular}
\end{footnotesize}
\end{flushleft}

\noindent
With $\bm{C}$ and $\bm{S}$ in hand, we can now proceed to set up the mappings from form to meaning (comprehension), and from meaning to form (production).  We first compute the comprehension network $\bm{F}$, which predicts word meanings from word forms, by solving $\bm{S} = \bm{CF}$.  {\bf JudiLing}'s function {\tt make\_transform\_matrix} estimates $\bm{F}$ in a numerically highly optimized way. 
\begin{jllisting}
F = JudiLing.make_transform_matrix(C_obj.C, S);
\end{jllisting}
The $\bm{F}$ matrix specifies the connection weights (in network parlance) or equivalently the beta weights (in the parlance of multivariate multiple regression) of all the di-syllable cues (rows) to all semantic dimensions (columns).  In other words, $\bm{F}$ is a trained network, or equivalently, a multivariate multiple regression model, that takes a verb's form vector $\bm{c}$ as input and predicts its semantic vector $\hat{\bm{s}}$. Restating this for all words jointly, we obtain the matrix with predicted semantic vectors $\hat{\bm{S}}$ by multiplying $\bm{C}$ with $\bm{F}$:
\begin{jllisting}
Shat = C_obj.C * F;
\end{jllisting}

\noindent
How successful is this mapping from form to meaning?  How close are the predicted semantic vectors to the gold standard vectors in $\bm{S}$?  We can evaluate closeness by means of the Pearson correlation, and evaluate the mapping to be successful for a given word $i$ if its predicted semantic vector $\hat{\bm{s}}_i$ has the highest correlation with the gold standard vector $\bm{s}_i$, that is, when
$$
r(\hat{\bm{s}}_i, \bm{s}_i) = \max_j\{r(\hat{\bm{s}}_j, \bm{s}_i)\}.
$$
The {\tt eval\_SC} function of {\bf JudiLing} evaluates model accuracy in this way.
\begin{jllisting}
JudiLing.eval_SC(Shat, S, korean, :Word)
0.9918
\end{jllisting}
For the present dataset, with over 27 thousand word forms, we obtain an accuracy of 99\%.

For modeling the production of Korean verb forms, we first derive the weight matrix $\bm{G}$, associated with the network that maps meanings $\bm{S}$ to forms $\bm{C}$. With $\bm{G}$ in hand, we calculate the predicted form matrix $\hat{\bm{C}}$.
\begin{jllisting}
G = JudiLing.make_transform_matrix(S, C_obj.C);
Chat = S * G;
\end{jllisting}
\noindent
The $\hat{\bm{C}}$ matrix specifies, for each word (rows) the amount of estimated semantic support that the di-syllables receive for that word. In the following tabulation, the di-syllable {\tt go\_rUm} is well-supported for the first and second words, but not for the third and fourth ones.
\begin{jllisting}
JudiLing.display_matrix(korean, :Word, C_obj, Chat, :Chat, nrow=4, ncol=7)
\end{jllisting}
\begin{flushleft}
\begin{footnotesize}
\begin{tabular}{l|lccccccc}\hline
	&  & \#\_go & go\_rUm & rUm\_ni & ni\_da & da\_\# & ni\_kka & kka\_\# \\
	\hline
	1 & go\_rUm\_ni\_da & 0.747749 & 0.184328 & 0.188502 & 0.656815 & 0.757723 & 0.190278 & 0.190721 \\
	2 & go\_rUm\_ni\_kka & 0.753465 & 0.185616 & 0.186941 & 0.211225 & 0.354903 & 0.541299 & 0.535272 \\
	3 & go\_rU\_sip-\_syo & 0.753545 & 0.027506 & 0.012114 & 0.225991 & 0.324489 & 0.223644 & 0.220002 \\
	4 & go\_rUp-\_si\_da & 0.753558 & 0.027182 & 0.031723 & 0.206195 & 0.728115 & 0.184292 & 0.185375 \\\hline
\end{tabular}
\end{footnotesize}
\end{flushleft}

\noindent
We now know which di-syllables are most likely to be part of a given word, but at this point, we have no criterion for selecting those di-syllables that actually occur in a given word, and we also have no information about their proper order. Since di-syllables contain partial information about syllable order, we can weave syllables together.  There are several ways in which this can be accomplished.  Here, we make use of an algorithm that first learns to predict, for a given word, and for all possible positions, which di-syllable is most appropriate for that position.  Given the top candidates, the algorithm then weaves together di-syllables into words, which typically results in several possible word candidates.  The algorithm then selects the candidate that, when presented to the comprehension network, best approximates the targeted meaning \citep[][refer to this as synthesis by analysis]{Baayen:Chuang:Shafei:Blevins:2019}.  The {\tt learn\_paths} function implements this algorithm. 
\begin{jllisting}
res = JudiLing.learn_paths(korean, C_obj, S, F, Chat);
\end{jllisting}

\noindent
We evaluate production accuracy by comparing the predicted forms with the targeted forms. Production accuracy is 90\%.
\begin{jllisting}
JudiLing.eval_acc(res, C_obj)
0.9006
\end{jllisting}

What we have shown, for a non-trivial Korean dataset of complex inflected verbs, is that the simple mappings of the Discriminative Lexicon model work with surprisingly high levels of accuracy for both production and for comprehension. It is beyond the scope of the present study to evaluate model accuracy on held-out data, for a first exploration of the productivity of LDL networks, see \citet{Chuang:Loo:Blevins:Baayen:2019}.  Now that we have shown that LDL can indeed learn the mappings between form and meaning without requiring stems, morphemes, or exponents, the question arises of whether the model can generate useful predictions for experimental data on lexical processing.  The next section takes on this question for comprehension.

\section{Comprehension}

\citet{creemers2020opacity} used the primed auditory lexical decision task to assess processing costs for Dutch complex verbs with initial particles or prefixes.  This study observed equivalent priming effects for semantically transparent and semantically opaque verbs, as compared to control conditions, replicating earlier work on German verbs \citep{smolka2007stem,Smolka:Eulitz:2018}.  These results dovetail well with theories positing that morphological processing involves automatic decomposition of forms into their constituents \citep{Taft:2004,stockall2006single,lewis2011neural,marantz2013no}, irrespective of their semantics. 

LDL does not decompose visual or auditory input into morphological constituents. Instead, it decomposes its input into finer-grained, overlapping sublexical units such as trigrams, triphones, or syllables \citep[for modeling audio input with LDL, see][]{ShafaeiBajestan:2021}.  Since the model sets up simple mappings between sublexical units and semantic vectors, it is a priori not self-evident that it can predict priming effects that suggest the presence of form effects in the absence of semantic effects.  \citet{Baayen:Smolka:2020} addressed this question for German, and showed, using the Na\"{i}ve Discriminative Learning (NDL) model of \citet{Baayen:Milin:Filipovic:Hendrix:Marelli:2011}, that priming effects are inversely proportional to the extent that the trigrams in a prime word pre-activate the semantic unit representing the meaning of the target word.  The downside of the NDL model is that all word meanings are  construed as orthogonal, thanks to its one-hot encoding of words' meanings, and thus completely unrelated semantically.  Hence, it might be argued that semantics does not really come into its own in this model, and that the NDL model of \citet{Baayen:Smolka:2020} is simply capturing pure form effects.  In what follows, we therefore move from NDL to LDL, and apply the latter model to the Dutch complex verbs used in Experiment~1 of \citet{creemers2020opacity}.  Importantly, unlike for our Korean dataset, where we simulated semantic vectors, in this modeling study we made use of empirical semantic vectors.  If LDL correctly predicts the observed priming effects, then this cannot be due to the model not having access to proper corpus-based word embeddings.

Input for modeling was a set of 7803 Dutch words, selected from the lemma database in CELEX \citep{Baayen:CD95}. For a word to be included, it had to have a frequency of occurrence exceeding 100 (per 42 million) and have at most two constituents, or it had to be listed as a prime or target in Experiment~1 of \citet{creemers2020opacity}. The resulting dataset is loaded into Julia as follows: 
\begin{jllisting}
dutch = DataFrame(CSV.File("dutch.csv"));
\end{jllisting}
For form vectors, we opted for using phone trigrams as sublexical units of form.
\begin{jllisting}
C_obj = JudiLing.make_cue_matrix(dutch, grams=3, target_col=:phon);
size(C_obj.C)
(7803, 6356)
\end{jllisting}
For the semantic matrix $\bm{S}$, we used the 300-dimensional vectors constructed with {\tt fasttext} as available at \url{https://fasttext.cc/docs/en/crawl-vectors.html}. The file {\tt S\_dutch.txt} contains the semantic vectors for the words in our dataset, in exactly the order in which they occur in {\tt dutch}. The first column of {\tt S\_dutch.txt} specifies the word, and the remaining 300 columns specify the word embedding. The vectors and the corresponding words are loaded into Julia as follows:
\begin{jllisting}
S, words = JudiLing.load_S_matrix("S_dutch.txt")
\end{jllisting} 
$\bm{S}$ is then the semantic matrix, and the object {\tt words} lists the words in the dataset. After obtaining the transformation matrix $\bm{F}$ and the predicted semantic vectors $\hat{\bm{S}}$,
\begin{jllisting}
F = JudiLing.make_transform_matrix(C_obj.C, S);  
Shat = C_obj.C * F; 
\end{jllisting}
we calculated all pairwise correlations of gold standard and predicted semantic vectors, and observed that for 91\% of words, the predicted semantic vector was the one that was closest to its targeted gold standard. The following code snippet calculates both the accuracy and the matrix of pairwise correlations ${\bm{R}}$ of words' estimated and gold standard semantic vectors, which we will need below. 
\begin{jllisting}
accuracy, R = JudiLing.eval_SC(Shat, S, dutch, :phon, R=true);
accuracy
0.909                     
\end{jllisting}

In order to model lexical priming, we conceptualize the reader's or listener's current semantic state as a pointer to a location in semantic space.  An experimental trial can then be seen as beginning with a semantic pointer at the origin of this space.  Presentation of the prime moves this pointer to the position in semantic space where the prime's semantic vector is approximately located.  We may expect that when this position is closer to that of the target word, the target word is primed more strongly.  We evaluated closeness with the correlation measure. For all prime-target pairs, we extracted from $\bm{R}$ the correlation $r_{ij}$ of the predicted vector for prime $i$ and the gold standard vector for target $j$.  In what follows, we refer to these correlations as prime-to-target approximations (PTA).  

\begin{figure}
    \centering
    \includegraphics[width=1.0\textwidth]{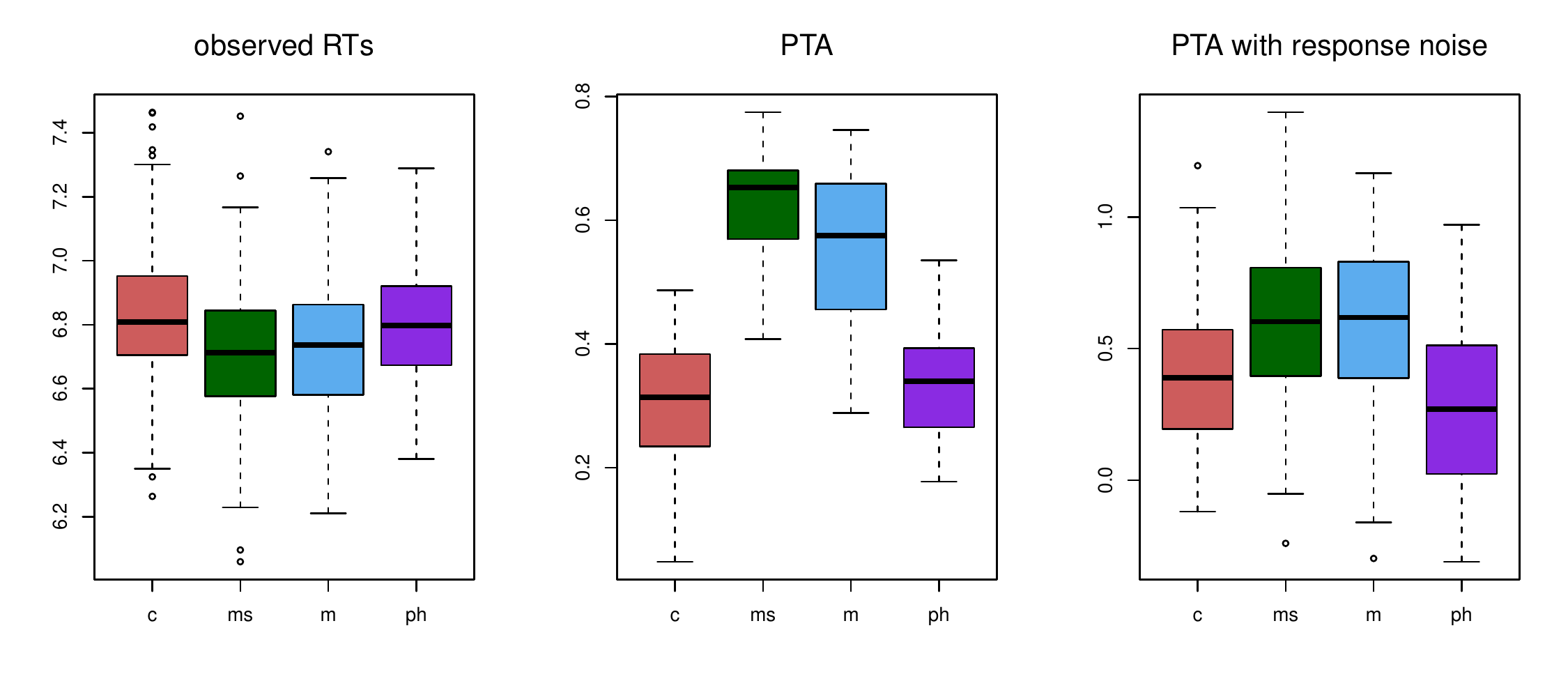}
    \vspace*{-2\baselineskip}
    \caption{Distributions of empirical reaction times (left), prime-to-target approximations (center) and prime-to-target approximations with ${\cal N}(0, 0.3)$ response noise added.}
    \label{fig:boxplots}
\end{figure}

Table~\ref{tab:dutch} and Figure~\ref{fig:boxplots} show that these correlations capture the priming effects reported by \citet{creemers2020opacity}.  As expected, greater values of PTA predict shorter RTs.  The correlation between PTA and mean by-item log-transformed reaction time \citep[as made available by][]{Creemers:2019}) was $-0.3$ ($t(141) = -3.67, p = 0.0003$). It is clear that LDL succeeds in modeling morphological priming data, replicating \citet{Baayen:Smolka:2020} \citep[see also][]{Milin:Feldman:Ramscar:Hendrix:Baayen:2017}.  It does so without ever having to parse forms into disjunct sets of morphological constituents.  

\begin{table}[ht]
\centering
\caption{Predicting prime-target approximation from prime type. The reference level for prime type is morphologically related but semantically unrelated prime (m); c: control primes, ph: phonologically related primes, ms: morphologically and semantically related primes.}
\vspace*{0.3\baselineskip}
\label{tab:dutch}
\begin{tabular}{lrrrr}\hline
 & Estimate & Std. Error & t value & Pr($>$$|$t$|$) \\ \hline
Intercept       & 0.5533 & 0.0166 & 33.40 & 0.0000 \\ 
  prime\_type = c  & -0.2464 & 0.0236 & -10.44 & 0.0000 \\ 
  prime\_type = ms &  0.0749 & 0.0234 & 3.20 & 0.0017 \\ 
  prime\_type = ph & -0.2187 & 0.0234 & -9.34 & 0.0000 \\ \hline
\end{tabular}
\end{table}

Having illustrated how measures derived from LDL can be used to predict lexical processing costs as gauged by a priming paradigm, in the next section we consider how the model can be used to predict words' acoustic durations.

\section{Speech production: Acoustic duration for Taiwan Mandarin words}

One measure which is informing theories of speech production is words' acoustic durations \citep{Gahl:2008,Gahl:Yao:Johnson:2012,Plag:Homann:Kunter:2017,Tomaschek:Plag:Ernestus:Baayen:2019}.  In this study, we investigated the acoustic durations of words in Taiwan Mandarin.  From a corpus of spontaneous Taiwan Mandarin speech \citep{fon2004}, we selected all 7,349 one- to five-syllable words for which {\tt fasttext} semantic vectors are available.  In the following snippet of Julia code, we assume the dataset {\tt mandarin.csv} is available in the working directory. 
\begin{jllisting}
mandarin = DataFrame(CSV.File("mandarin.csv"));
\end{jllisting}
An overview of this dataset is presented in Table~\ref{tab:man_ex}.

\begin{table}[h]
\caption{The first four rows of the Mandarin dataset. Glosses for the words from top to bottom: `one', `one by one', `a bit', and `abruptly'.}
\centering
\begin{tabular}{r|lll} \hline
	  & word                & pinyin         & phones         \\	\hline
	1 & \mandarin{一}       & yi1            & i1             \\
	2 & \mandarin{一一}     & yi1\_yi1       & i1.i1          \\
	3 & \mandarin{一下}     & yi1\_xia4      & i1.x.ia4       \\
	4 & \mandarin{一下子}   & yi1\_xia4\_zi5 & i1.x.ia4.z.ii5 \\ \hline
\end{tabular}
\label{tab:man_ex}
\end{table}

\noindent
In this dataset, the second column ({\tt pinyin}) provides the transcription of the words according to the standard romanization system. For modeling, however, we used the more detailed transcription of phones in the third column ({\tt phones}). Here, phones are separated by periods, and Mandarin lexical tones are indicated by numbers next to the vowels.  As in the previous study on Dutch,  we used triphones to build form vectors.
\begin{jllisting}
C_obj = JudiLing.make_cue_matrix(mandarin, grams=3, target_col=:phones,
                                 tokenized=true, sep_token=".", keep_sep=true);
\end{jllisting}
Given a file with {\tt fasttext} vectors ({\tt S\_mandarin.txt}), with words' vectors listed in the same order as the words in the dataset of {\tt mandarin}, we load the semantic matrix into Julia,
\begin{jllisting}
S, words = JudiLing.load_S_matrix("S_mandarin.txt");
\end{jllisting}
and then calculate the transformation matrix $\bm{F}$, the matrix with predicted semantic vectors $\hat{\bm{S}}$ :
\begin{jllisting}
F = JudiLing.make_transform_matrix(C_obj.C, S);
Shat = C_obj.C * F;
\end{jllisting}
Comprehension accuracy is high:
\begin{jllisting}
JudiLing.eval_SC(Shat, S, mandarin, :phones)
0.9877
\end{jllisting}
For modeling production, we first estimate the mapping $\bm{G}$ and the matrix with predicted triphone supports $\hat{\bm{C}}$,
\begin{jllisting}
G = JudiLing.make_transform_matrix(S, C_obj.C);
Chat = S * G;
\end{jllisting}
and then use the {\tt learn\_paths} function to weave the triphones into words. 
\begin{jllisting}
res, gpi = JudiLing.learn_paths(mandarin, C_obj, S, F, Chat, threshold=0.01,
                                Shat_val=Shat, check_gold_path=true);
\end{jllisting}
The directive {\tt check\_gold\_path}, when set to {\tt true}, requests the function to keep track of the amount of support that the triphones of a given word receive in different positions, and outputs a second object {\tt gpi}. As will become clearer later, we will need this object to derive network measures.  We note here that for production, accuracy is also satisfactory:
\begin{jllisting}
JudiLing.eval_acc(res, C_obj)
0.9291
\end{jllisting}
\noindent
From this model of the Mandarin mental lexicon, we derived two novel measures that are predictive for Mandarin words' acoustic durations.

The first measure is derived from the production network.  Recall that the weight matrix of this network, $\bm{G}$, has semantic dimensions on rows and triphones on the columns.  The column vectors of $\bm{G}$ quantify the extent to which triphones contribute to realizing the meanings of their carrier words. To make this more concrete, consider an example lexicon with just four English words: {\em tree}, {\em bee}, and their plural forms {\em trees}, {\em bees}. Model setup is presented in Table~\ref{tab:toy_ex}: for cues we used diphones, and their semantic vectors, using simulated vectors, are the sum of their respective base word vector ({\sc tree}/{\sc bee}) and number vector ({\sc sg}/{\sc pl}). Following the same procedure as described above, we obtained the $\bm{G}$ matrix, the production network mapping meanings onto forms. Table~\ref{tab:dist_ex} presents the column vectors of $\bm{G}$, once for {\em trees} (left) and once for {\em bees} (right), together with a vector of zeroes representing the origin.

\begin{table}[h]
\centering
\caption{Example lexicon with four words. The second and third columns specify words' diphones and how their semantic vectors are constructed.}
\begin{tabular}{lll}
\hline
Word & diphones          & semantics \\ \hline
tree & \#t tr ri i\#     & $\overrightarrow{\text{TREE}} + \overrightarrow{\text{SG}}$ \\
bee  & \#b bi i\#        & $\overrightarrow{\text{BEE}} + \overrightarrow{\text{SG}}$ \\
trees& \#t tr ri iz z\#  & $\overrightarrow{\text{TREE}} + \overrightarrow{\text{PL}}$ \\
bees & \#b bi iz z\#     & $\overrightarrow{\text{BEE}} + \overrightarrow{\text{PL}}$ \\
\hline        
\end{tabular}
\label{tab:toy_ex}
\end{table}

\begin{figure}
\centering
\includegraphics[width=0.5\textwidth]{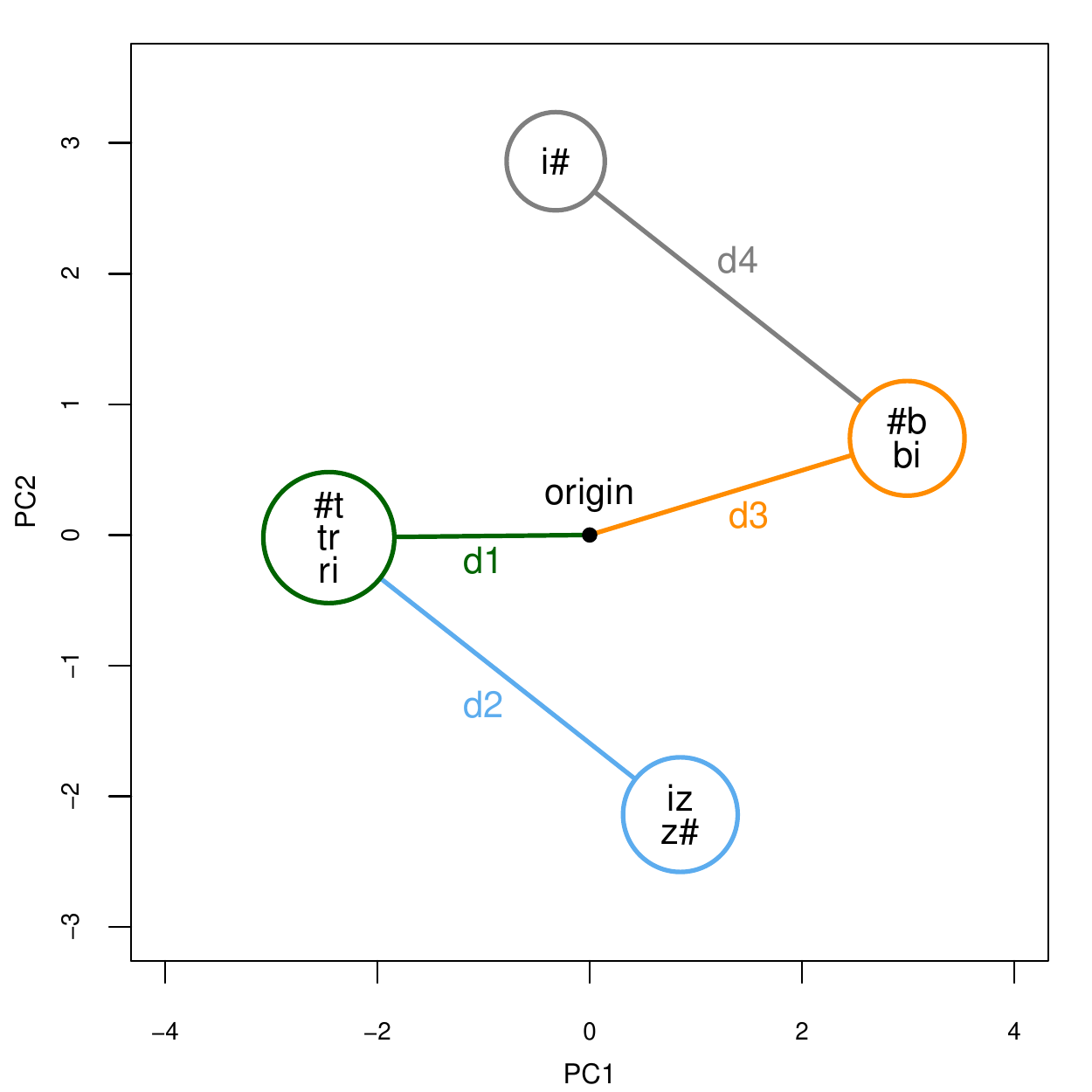}
\caption{Projection of diphone vectors onto a two-dimensional `cortical' plane, using principal component analysis. For {\em trees}, the total Distance Travelled is d1 + 0 + 0 + d2 + 0. For {\em bee}, the distance is d3 + 0 + d4 (see also Table~\ref{tab:dist_ex}).}
\label{fig:pointer}
\end{figure}

From a neurobiological perspective, we can regard the $\bm{G}$ matrix, in the roughest sense, as a production neural network, with semantic features and diphone cues as representational units of neurons (or collections of neurons) in the cortex. As neurons have a location in the cortex, we therefore also examine what a sensible topological organization of the diphone units of  our network would be. Note that the order in which diphones are listed in the columns of the $\bm{G}$ matrix is completely arbitrary --- all that counts for the mathematics is their identity and their connectivity.  By contrast, neurons in the cortex have non-arbitrary topologies \citep[see, e.g.,][]{ferro2011self}. There are many different ways of projecting representations in a high-dimensional space to a two-dimensional surface that could serve as a rough approximation of a (highly simplified) cortical surface \citep[see also][]{Heitmeier:Baayen:2021,Baayen:Chuang:Blevins:2018}. Here we approximated such a surface by applying principal component analysis to the column vectors of $\bm{G}$. As can be seen from Figure~\ref{fig:pointer}, the stem diphones of {\em tree} ({\tt \#t}, {\tt tr}, {\tt ri}) appear together on the left, marked by green circle, whereas the stem diphones of {\em bee} ({\tt \#b}, {\tt bi}) cluster in the orange circle on the right. Interestingly, the diphone {\tt i\#} on the one hand, and those of {\tt iz} and {\tt z\#} on the other, also are located in two different positions, at the top and bottom respectively.

\begin{table}
\caption{Cue weights in the production network $\bm{G}$. Table (a) and (b) present diphones required to produce {\em trees} and {\em bee}, respectively. The bottom row indicates pair-wise Euclidean distances between diphone vectors, starting from the origin (all-zero vector) to the last diphone of the word.}
\begin{subtable}{.48\linewidth}
\centering
\caption{{\em trees}}
\begin{small}
\begin{tabular}{c|c|rrrrr} \hline
         & origin &   \#t&    tr &   ri &     iz &    z\# \\ \hline
S1       &      0 &-0.01 & -0.01 &-0.01 &  -0.03 & -0.03  \\
S2       &      0 &-0.06 & -0.06 &-0.06 &  -0.11 & -0.11  \\
S3       &      0 & 0.05 &  0.05 & 0.05 &  -0.04 & -0.04  \\
S4       &      0 & 0.07 &  0.07 & 0.07 &   0.03 &  0.03  \\
S5       &      0 &-0.02 & -0.02 &-0.02 &   0.01 &  0.01  \\
S6       &      0 & 0.00 &  0.00 & 0.00 &  -0.02 & -0.02  \\
S7       &      0 &-0.07 & -0.07 &-0.07 &   0.00 &  0.00  \\
S8       &      0 &-0.05 & -0.05 &-0.05 &  -0.01 & -0.01  \\ \hline
\multicolumn{1}{c|}{distance} &    
  \multicolumn{6}{c}{\begin{tabular}[t]{p{0.8cm}p{0.8cm}p{0.8cm}p{0.8cm}c} 
      \hspace*{-0.5em}0.137 & \hspace*{0.5em}0 & \hspace*{0.8em}0 & 0.143 & \hspace*{0.5em}0  
     \end{tabular}}\\ \hline
\end{tabular}
\end{small}
\end{subtable}
\hfill
\begin{subtable}{.48\linewidth}
\centering
\caption{{\em bee}}
\begin{small}
\begin{tabular}{c|rrrr} \hline
 origin &   \#b&    bi &     i\# \\ \hline
      0 &-0.01 &-0.01 &   0.01   \\
      0 & 0.04 & 0.04 &   0.08   \\
      0 &-0.11 &-0.11 &  -0.02   \\
      0 &-0.03 &-0.03 &   0.00   \\
      0 & 0.07 & 0.07 &   0.03   \\
      0 &-0.02 &-0.02 &   0.00   \\
      0 & 0.04 & 0.04 &  -0.03   \\
      0 & 0.00 & 0.00 &  -0.05   \\ \hline
\multicolumn{4}{c}{\begin{tabular}[t]{p{0.8cm}p{0.8cm}p{0.8cm}c} 
      \hspace*{-0.1em}0.147 & \hspace*{0.5em}0 & 0.143   
     \end{tabular}}\\ \hline
\end{tabular}
\end{small}
\end{subtable}
\label{tab:dist_ex}
\end{table}

To understand why the diphones cluster in this way, we have to go back the column vectors of $\bm{G}$ (henceforth diphone vectors). Since the diphone vectors share the same dimensions as the semantic vectors, we can examine how they are semantically related by calculating the pair-wise correlations of the diphone vectors and the vectors of all base words and inflectional meanings --- {\sc tree, bee, sg, pl}, referred to as lexomes in \citet{Baayen:Chuang:Shafei:Blevins:2019}. The resulting correlation matrix is presented in Table~\ref{tab:toy_cor}. The diphones of {\em tree} are best correlated with $\overrightarrow{\text{TREE}}$, and the diphones of {\em bee} are best correlated with $\overrightarrow{\text{BEE}}$. Furthermore, {\tt i\#} is strongly correlated with $\overrightarrow{\text{SG}}$, whereas {\tt iz} and {\tt z\#} correlate well with $\overrightarrow{\text{PL}}$. The correlations in bold in Table~\ref{tab:toy_cor} can therefore be understood as measuring the functional load of the diphones. In this example, {\tt i\#} realizes primarily singular number, {\tt z\#} realizes plural number, and {\tt \#b} and {\tt bi} realize $\overrightarrow{\text{BEE}}$.\footnote{These correlations clarify how those diphones are selected for production that properly realize words' semantics.  When we calculate the dot product of a row vector $\bm{s}_i$ of $\bm{S}$ (the sum of two lexome vectors, e.g., $\overrightarrow{TREE}$ + $\overrightarrow{PL}$) and a diphone column vector $\bm{g}_j$ of $\bm{G}$, we multiply their elements pairwise and then sum. When the two vectors are uncorrelated, this sum will be close to zero. However, when the two vectors are correlated, the dot product will tend to produce larger values.  As a consequence, correlated vectors ensure that the support for a diphone in a word's predicted form vector $\hat{\bm{c}}$ will be greater.} It is surprising how well the network succeeds in figuring out, for this agglutinative example, which bits of form have what purpose, without ever having to tell the model explicitly about stems and exponents.\footnote{For the present agglutinative example, we see clusters of diphones for the two stems and the plural exponent, but the reader should be warned that for realistically sized datasets, and especially for languages with fusional morphology, such crisp and clear divisions of labor are not to be expected.  Nevertheless, some albeit more diffuse clustering may still be visible.}

\begin{table}[]
    \centering
    \caption{Correlations of the diphone vectors in $\bm{G}$ and the semantic vectors of the base word and number lexomes.  The highest correlation in each row is highlighted in bold.}
    \begin{tabular}{lrrrr}
     \hline
                    &  {\sc tree}  & {\sc bee}  & {\sc sg}  & {\sc pl}  \\ \hline
      \#t tr ri     &  {\bf 0.55}  &   -0.44    &   0.26    &    0.13   \\
      \#b bi        &   -0.42      & {\bf 0.72} &   0.15    &    0.28   \\ 
      i\#           &    0.04      &    0.19    &{\bf 0.85} &   -0.17   \\ 
      iz z\#        &    0.11      &    0.18    &  -0.29    &{\bf 0.68} \\ \hline
    \end{tabular}
    \label{tab:toy_cor}
\end{table}

Is lexical processing, specifically durations of word production in this study, influenced by the proximity of diphones in such a hypothesized cortical 2-D plane? We reasoned that when diphones have very similar functional load, the costs of evaluating their contribution to production is smaller than when diphones are at greater distance of each other.  This hypothesis leads to our first measure for predicting acoustic durations, the total distance travelled in semantic space when tracing a word's successive diphones (henceforth {\tt Distance Travelled}).  Consider again Table~\ref{tab:dist_ex}(a), which arranges the diphones in the order in which they drive articulation.  Assuming that we start from the null meaning (represented by the origin, an all-zero vector), we can calculate the Euclidean distance from one successive diphone vector to the next, as indicated in the last row of the table. From the origin to the vector of {\tt \#t}, this distance is 0.137. In Figure~\ref{fig:pointer}, this distance is labeled as {\bf d1}, in green.   From {\tt \#t} to {\tt tr}, and from {\tt tr} to {\tt ri}, the distance is zero:  these three diphones are associated with the {\sc tree} meaning to the same extent, and have the same location in Figure~\ref{fig:pointer}.  Hence for these diphones, no further distance penalty is incurred. However, for the next transition, as {\tt iz} is located further down in the hypothesized cortical plane, we obtain a non-zero distance ({\bf d2} in Figure~\ref{fig:pointer}). Finally, {\tt z\#} is in the same position as {\tt iz}, hence the distance between the last two diphones is zero.  The total Distance Travelled for {\em trees} is 0.137 (d1) + 0 + 0 + 0.143 (d2) + 0 = 0.28. For {\em bee} (Table~\ref{tab:dist_ex}(b)), the total Distance Travelled is 0.147 (d3) + 0 + 0.143 (d4) = 0.29.  (For more realistic datasets, distances between diphones or triphones will rarely be zero, as the network weights for transitions represent compromises between many conflicting constraints.) The larger the value of Distance Travelled, the more heterogeneous the diphone or triphone vectors vectors are, the more variegated the meanings are that they have to realize, and the lower the within-word semantic coherence is.   These considerations lead us to expect that acoustic duration and Distance Travelled are positively correlated: more complex messages require longer codes \citep[cf.][]{Kuperman:Schreuder:Bertram:Baayen:2009}.

The second measure that we considered for predicting acoustic durations is much simpler. As illustrated in Section~\ref{sec:korean}, the learning algorithm that orders triphones for articulation (implemented in the {\tt learn\_paths} function) predicts, for a given word and a specific triphone position (ranging from 1 to the length of the longest word in the dataset), how well the different possible triphones are supported for that position.  For example, for the Mandarin word \mandarin{土地} (t\v{u}d\`{i}) `land', the positional triphone support for the first triphone {\tt \#tu} is 0.04 for position 1, and that for the second triphone {\tt tud} is 0.03 for position 2.  Our second measure, {\tt Triphone Support}, is the sum of the positional supports for a words' diphones. The larger the value of Triphone Support, the more confident the model is in terms of its prediction. We therefore expect that Triphone Support and acoustic duration enter into a positive correlation. 

These two measures can be straightforwardly obtained with the {\bf JudiLing} package: 
\begin{jllisting}
distances = JudiLing.get_total_distance(C_obj, G, :G);
supports = JudiLing.get_total_support(gpi);
\end{jllisting}
Both functions return vectors with the pertinent values for each of the words in the dataset.

Are these two measures indeed predictive for words' acoustic durations?  We first fitted a generalized additive model (GAM) \citep{Wood:2017} to the acoustic durations with word frequency and word length (number of phones) as predictors. Both frequency and length are strong predictors, unsurprisingly. The effect of word length is rather trivial, as all we are doing is regressing duration in milliseconds on duration in phones.  The reason we include length is to ensure that our model-based measures cannot simply be reduced to word length.  In other words, this model serves as our baseline model.

\begin{figure}
\centering
\includegraphics[width=0.98\textwidth]{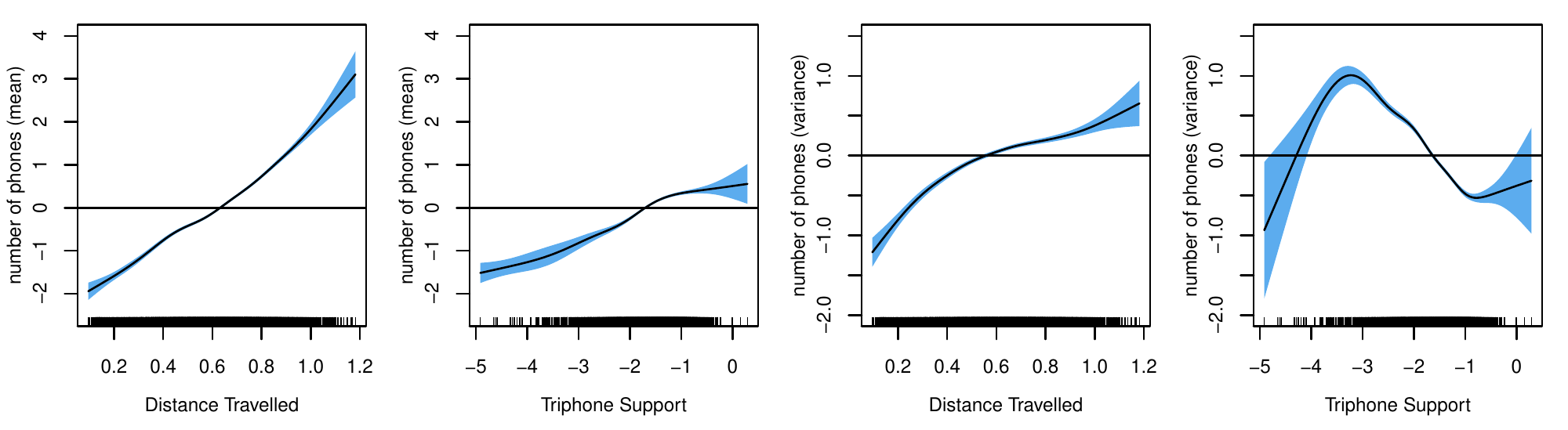}
\caption{The effects of Distance Travelled and Triphone Support on word length (number of phones). The left two panels show that mean word length increases with both predictors. The right two panels show that the variability in word length increases with Distance Travelled, but decreases with Triphone Support.}
\label{fig:man_length}
\end{figure}

When Distance Travelled and Triphone Support are added as predictors to the baseline model, goodness of fit improved substantially by no less than than 195 AIC units. This clarifies that the distance and support measures capture properties in production that cannot be fully accounted for by word length and frequency.   Unfortunately, due to substantial collinearity, the resulting model suffers from suppression and enhancement, rendering interpretation impossible.  Both our model-based measures are  highly correlated with word length: for Distance Travelled, $r = 0.81$, $p < 0.0001$, and for Triphone Support, $r = 0.37$, $p < 0.0001$.  In order to obtain a model that is better interpretable,  we first regressed word length on the distance and support measures. This was achieved by fitting another GAM to word length with the two LDL measures as predictors. Instead of a standard GAM, we used a Gaussian location-scale model, which models both mean and variance of word length at the same time.  The two  measures are predictive for both the mean and the variance of length (see Figure~\ref{fig:man_length}). In general, longer words have longer distances travelled in semantic space, as well as larger support received by triphones (left two panels). Interestingly, the variance of word length also varies with the two LDL measures. While variability in length increases with Distance Travelled, it decreases with Triphone Support.\footnote{Interestingly, once the two LDL measures are in the model, frequency of occurrence is no longer predictive for word length.}  The R-squared of this model was 49\%. The correlation of the residuals of this model, henceforth {\tt Residualized Length}, with the original word length measure was 0.72.

Whereas the positive correlations of the two measures with mean length are unsurprising, their strong effect on the variance in word length is novel and of potential theoretical interest.   For Triphone Support, greater support offers more precise prediction of word length (as variance in word length decreases). In other words, when the model struggles with learning triphone positions, resulting in low values, predictions for word length are more variable.  By contrast, for larger values of Distance Travelled, predictions for word length apparently become less precise.

\begin{figure}[b]
\centering
\includegraphics[width=0.95\textwidth]{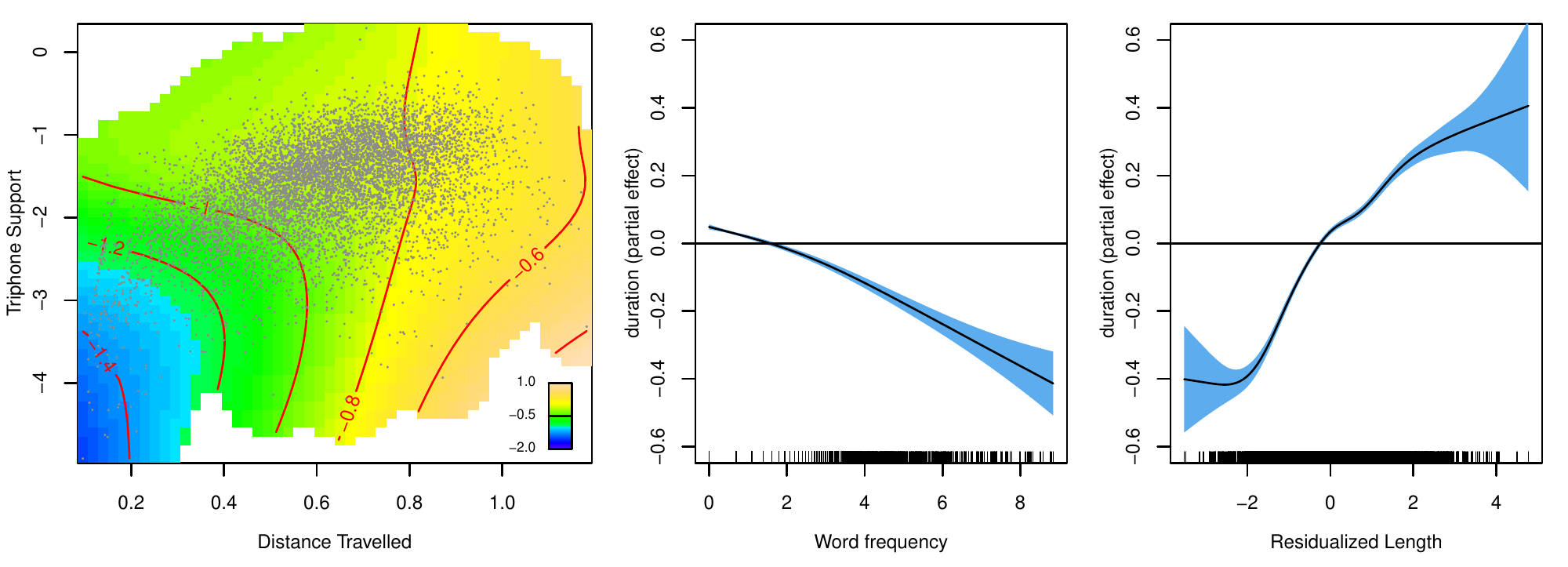}
\caption{The interaction effect of Distance Travelled and Triphone Support on mean acoustic durations of Mandarin words produced in spontaneous speech (left panel). Warmer colors represent longer durations, and dots indicate the distribution of data points. The effects of word frequency, and Residualized Length are presented in the middle and right panels.}
\label{fig:man_dur}
\end{figure}

We then re-fitted the GAM, replacing word length with Residualized Length. The mid and right panels of Figure~\ref{fig:man_dur} plot the effects of frequency and length. Words with higher frequency have shorter durations, and phonologically longer words are produced with longer durations, as expected. What is more interesting is the interaction of the two LDL measures, as presented in the left panel. We observed a strong effect of Distance Travelled: word durations become longer with increasing  Distance Travelled.  This suggests that when a word's triphones are far apart in semantic space, and possibly far apart in our hypothetical cortical plane, in which case they jointly realize more complex semantics, then the speed with which this word is articulated is slowed, reflecting greater processing costs.   With regards to Triphone Support, there is a cross-over effect. For small values of Distance Travelled, the larger the support, the longer the durations. For large values of Distance Travelled, however, the trend is reversed, here durations become shorter with increasing Triphone Support.

In summary, with empirical semantic vectors, LDL performed well for both comprehending and producing Mandarin words. Furthermore, when pitted against behavioral data, the measures derived from the networks are predictive for the acoustic durations of these words as actually pronounced in spontaneous speech. The results suggest that the LDL framework, despite making use of very simple networks, is useful for studying lexical processing.

\section{General Discussion}

The three case studies that we have reported illustrate why we find it fruitful to rethink morphology in terms of mappings between form and meaning spaces, using the toolkit of linear algebra and regression modeling.  Although firmly rooted in Word and Paradigm Morphology, our analysis of how triphones may cluster in a two-dimensional approximation of a cortical plane suggests that, when a morphological system is agglutinative, clusters of triphones start to look very much like stems and exponents.  If this finding generalizes from a toy example to realistic datasets, it would provide a bridge between Word and Paradigm Morphology and decompositional theories that are supported by neuroimaging studies suggesting that constituents are subserved by localized cell-assemblies in the brain \citep{bozic2007differentiating,lewis2011neural,chersi2014topological}. At the same time, it is clear that for claims about the neurobiological reality of classical discrete morpheme and stem representations to be convincing, it will be necessary to show that there is no confound between the classical units and sublexical units such as triphones.  Likewise, when designing priming studies to support the hypothesis of obligatory automatic morphological decomposition, it will be necessary to control for the kind of low-level variables that this study has shown to co-determine lexical processing, and that patterns of priming do not already emerge straightforwardly from a language's distributional properties.  

\citet{breiman2001statistical} famously made a distinction between two very different cultures for data analysis, contrasting statistical modeling with machine learning. Whereas machine learning is optimized for obtaining precise predictions that can be profitably put to use in industry, statistical modeling is concerned with understanding the data, and to this end builds models that could have generated the data.  LDL is closer in spirit to statistical modeling than to machine learning.  It capitalizes on simplicity, in the hope that this will further clarity of understanding.  Undoubtedly, this simplicity comes at a cost.  Although many mappings in morphology thus far have appeared to be well approximated by linear mappings, nonlinear mappings are likely to offer further precision.  Interestingly, exactly the words for which LDL lacks precision are the words that are likely to be more difficult to learn also for human learners, and that hence will have greater lexical processing costs in the mental lexicon.  George Box is well known for stating that all statistical models are wrong, but some are useful \citep{Box:1976}.  In the same way, even though vector space morphology as implemented with LDL requires many simplifying assumptions, it nevertheless provides a useful tool for probing the mental lexicon.

\bibliography{vec_morphology}

\begin{thebibliography}{}

\bibitem[Baayen et~al., 2018]{Baayen:Chuang:Blevins:2018}
Baayen, R.~H., Chuang, Y.-Y., and Blevins, J.~P. (2018).
\newblock Inflectional morphology with linear mappings.
\newblock {\em The Mental Lexicon}, 13(2):232--270.

\bibitem[Baayen et~al., 2019]{Baayen:Chuang:Shafei:Blevins:2019}
Baayen, R.~H., Chuang, Y.-Y., Shafaei-Bajestan, E., and Blevins, J. (2019).
\newblock The discriminative lexicon: {A} unified computational model for the
  lexicon and lexical processing in comprehension and production grounded not
  in (de)composition but in linear discriminative learning.
\newblock {\em Complexity}.

\bibitem[Baayen et~al., 2011]{Baayen:Milin:Filipovic:Hendrix:Marelli:2011}
Baayen, R.~H., Milin, P., Filipovi\'c~Durdevi\'c, D., Hendrix, P., and Marelli,
  M. (2011).
\newblock An amorphous model for morphological processing in visual
  comprehension based on naive discriminative learning.
\newblock {\em {Psychological Review}}, 118:438--482.

\bibitem[Baayen et~al., 1995]{Baayen:CD95}
Baayen, R.~H., Piepenbrock, R., and Gulikers, L. (1995).
\newblock {\em The {CELEX} lexical database (CD-ROM)}.
\newblock Linguistic Data Consortium, University of Pennsylvania, Philadelphia,
  PA.

\bibitem[Baayen and Smolka, 2020]{Baayen:Smolka:2020}
Baayen, R.~H. and Smolka, E. (2020).
\newblock Modelling morphological priming in {G}erman with naive discriminative
  learning.
\newblock {\em {Frontiers in Communication, section Language Sciences}}.
\newblock preprint on PsyArXiv, doi:10.31234/osf.io/nj39v.

\bibitem[Box, 1976]{Box:1976}
Box, G. E.~P. (1976).
\newblock Science and statistics.
\newblock {\em Journal of the American Statistical Association}, 71:791--799.

\bibitem[Bozic et~al., 2007]{bozic2007differentiating}
Bozic, M., Marslen-Wilson, W.~D., Stamatakis, E.~A., Davis, M.~H., and Tyler,
  L.~K. (2007).
\newblock Differentiating morphology, form, and meaning: Neural correlates of
  morphological complexity.
\newblock {\em Journal of cognitive neuroscience}, 19(9):1464--1475.

\bibitem[Breiman et~al., 2001]{breiman2001statistical}
Breiman, L. et~al. (2001).
\newblock Statistical modeling: The two cultures (with comments and a rejoinder
  by the author).
\newblock {\em {Statistical Science}}, 16(3):199--231.

\bibitem[Chersi et~al., 2014]{chersi2014topological}
Chersi, F., Ferro, M., Pezzulo, G., and Pirrelli, V. (2014).
\newblock Topological self-organization and prediction learning support both
  action and lexical chains in the brain.
\newblock {\em Topics in cognitive science}, 6(3):476--491.

\bibitem[Choi, 1929]{choi1929uli}
Choi, H.-B. (1929).
\newblock {Uli Mal Bon} {(Korean Grammar)}.
\newblock {\em Seoul: Chom-Um-5a, 6th edition.}

\bibitem[Chuang and Baayen, 2021]{Chuang:Baayen:2021}
Chuang, Y.~Y. and Baayen, R.~H. (2021).
\newblock Discriminative learning and the lexicon: {NDL} and {LDL}.
\newblock In {\em Oxford Research Encyclopedia of Linguistics}. {Oxford
  University Press} (to appear).

\bibitem[Chuang et~al., 2020a]{Chuang:Bell:Banke:Baayen:2020}
Chuang, Y.-Y., Bell, M.~J., Banke, I., and Baayen, R.~H. (2020a).
\newblock Bilingual and multilingual mental lexicon: a modeling study with
  linear discriminative learning.
\newblock {\em Language Learning}, pages 1--73.

\bibitem[Chuang et~al., 2020b]{Chuang:Loo:Blevins:Baayen:2019}
Chuang, Y.-Y., Loo, K., Blevins, J.~P., and Baayen, R.~H. (2020b).
\newblock {Estonian case inflection made simple. A case study in Word and
  Paradigm morphology with Linear Discriminative Learning.}
\newblock In K\"{o}rtv\'{e}lyessy, L. and \v{S}tekauer, P., editors, {\em
  Advances in Morphology}, pages 119--141. {Cambridge University Press}.

\bibitem[Creemers, 2019]{Creemers:2019}
Creemers, A. (2019).
\newblock Data for: Opacity, transparency, and morphological priming: A study
  of prefixed verbs in {D}utch.

\bibitem[Creemers et~al., 2020]{creemers2020opacity}
Creemers, A., Davies, A.~G., Wilder, R.~J., Tamminga, M., and Embick, D.
  (2020).
\newblock Opacity, transparency, and morphological priming: A study of prefixed
  verbs in dutch.
\newblock {\em Journal of Memory and Language}, 110:104055.

\bibitem[Ferro et~al., 2011]{ferro2011self}
Ferro, M., Marzi, C., and Pirrelli, V. (2011).
\newblock A self-organizing model of word storage and processing: implications
  for morphology learning.
\newblock {\em Lingue e linguaggio}, 10(2):209--226.

\bibitem[Fon, 2004]{fon2004}
Fon, J. (2004).
\newblock A preliminary construction of {T}aiwan {S}outhern {M}in spontaneous
  speech corpus.
\newblock Technical Report NSC-92-2411-H-003-050-, National Science Council,
  {Taipei, Taiwan}.

\bibitem[Gahl, 2008]{Gahl:2008}
Gahl, S. (2008).
\newblock Time and thyme are not homophones: The effect of lemma frequency on
  word durations in spontaneous speech.
\newblock {\em Language}, 84(3):474--496.

\bibitem[Gahl et~al., 2012]{Gahl:Yao:Johnson:2012}
Gahl, S., Yao, Y., and Johnson, K. (2012).
\newblock Why reduce? phonological neighborhood density and phonetic reduction
  in spontaneous speech.
\newblock {\em Journal of Memory and Language}, 66(4):789--806.

\bibitem[Heitmeier and Baayen, 2020]{Heitmeier:Baayen:2021}
Heitmeier, M. and Baayen, R.~H. (2020).
\newblock {Simulating phonological and semantic impairment of English tense
  inflection with Linear Discriminative Learning}.
\newblock {\em {The Mental Lexicon}}, 15(3):384--421.

\bibitem[Kuperman et~al., 2009]{Kuperman:Schreuder:Bertram:Baayen:2009}
Kuperman, V., Schreuder, R., Bertram, R., and Baayen, R.~H. (2009).
\newblock Reading of multimorphemic {D}utch compounds: {T}owards a multiple
  route model of lexical processing.
\newblock {\em Journal of {E}xperimental {P}sychology: {HPP}}, 35:876--895.

\bibitem[Landauer and Dumais, 1997]{Landauer:Dumais:1997}
Landauer, T. and Dumais, S. (1997).
\newblock A solution to {P}lato's problem: {T}he latent semantic analysis
  theory of acquisition, induction and representation of knowledge.
\newblock {\em Psychological {R}eview}, 104(2):211--240.

\bibitem[Lee, 1989]{lee1989korean}
Lee, H.~B. (1989).
\newblock {\em Korean grammar}.
\newblock Oxford University Press, New York.

\bibitem[Lewis et~al., 2011]{lewis2011neural}
Lewis, G., Solomyak, O., and Marantz, A. (2011).
\newblock The neural basis of obligatory decomposition of suffixed words.
\newblock {\em Brain and language}, 118(3):118.

\bibitem[Luo et~al., 2021]{Luo:Chuang:Baayen:2021}
Luo, X., Chuang, Y.-Y., and Baayen, R.~H. (2021).
\newblock Judiling: an implementation in {J}ulia of {Linear Discriminative
  Learning} algorithms for language modeling.
\newblock \url{https://megamindhenry.github.io/JudiLing.jl/stable/}.

\bibitem[Marantz, 2013]{marantz2013no}
Marantz, A. (2013).
\newblock No escape from morphemes in morphological processing.
\newblock {\em Language and Cognitive Processes}, 28(7):905--916.

\bibitem[Martin, 1960]{martin1960korean}
Martin, S.~E. (1960).
\newblock {\em Korean reference grammar}.
\newblock American Council of Learned Societies.

\bibitem[Mikolov et~al., 2013]{mikolov2013distributed}
Mikolov, T., Sutskever, I., Chen, K., Corrado, G.~S., and Dean, J. (2013).
\newblock Distributed representations of words and phrases and their
  compositionality.
\newblock In {\em Advances in neural information processing systems}, pages
  3111--3119.

\bibitem[Milin et~al., 2017]{Milin:Feldman:Ramscar:Hendrix:Baayen:2017}
Milin, P., Feldman, L.~B., Ramscar, M., Hendrix, P., and Baayen, R.~H. (2017).
\newblock Discrimination in lexical decision.
\newblock {\em {PLOS-one}}, 12(2):e0171935.

\bibitem[Mitchell and Lapata, 2008]{Mitchell:Lapata:2008}
Mitchell, J. and Lapata, M. (2008).
\newblock Vector-based models of semantic composition.
\newblock In {\em ACL}, pages 236--244.

\bibitem[Plag et~al., 2017]{Plag:Homann:Kunter:2017}
Plag, I., Homann, J., and Kunter, G. (2017).
\newblock Homophony and morphology: {T}he acoustics of word-final {S} in
  {E}nglish.
\newblock {\em Journal of {L}inguistics}, 53(1):181--216.

\bibitem[Shafaei-Bajestan et~al., 2021]{ShafaeiBajestan:2021}
Shafaei-Bajestan, E., Tari, M.~M., and Baayen, R.~H. (2021).
\newblock {LDL-AURIS}: Error-driven learning in modeling spoken word
  recognition.
\newblock {\em Language, Cognition and Neuroscience}.

\bibitem[Smolka and Eulitz, 2018]{Smolka:Eulitz:2018}
Smolka, E. and Eulitz, C. (2018).
\newblock Psycholinguistic measures for german verb pairs: Semantic
  trans-parency, semantic relatedness, verb family size, and age of reading
  acquisition.
\newblock {\em {Behavior Research Methods}}, 50(4):1540--1562.

\bibitem[Smolka et~al., 2007]{smolka2007stem}
Smolka, E., Zwitserlood, P., and R{\"o}sler, F. (2007).
\newblock Stem access in regular and irregular inflection: Evidence from
  {G}erman participles.
\newblock {\em {Journal of Memory and Language}}, 57(3):325--347.

\bibitem[Sohn, 2013]{sohn2013korean}
Sohn, H.~M. (2013).
\newblock {\em Korean}, volume~4.
\newblock Korea University Press.

\bibitem[Stockall and Marantz, 2006]{stockall2006single}
Stockall, L. and Marantz, A. (2006).
\newblock A single route, full decomposition model of morphological complexity:
  {MEG} evidence.
\newblock {\em {The Mental Lexicon}}, 1(1):85--123.

\bibitem[Taft, 2004]{Taft:2004}
Taft, M. (2004).
\newblock Morphological decomposition and the reverse base frequency effect.
\newblock {\em {The Quarterly Journal of Experimental Psychology}},
  57A:745--765.

\bibitem[Tomaschek et~al., 2019]{Tomaschek:Plag:Ernestus:Baayen:2019}
Tomaschek, F., Plag, I., Ernestus, M., and Baayen, R.~H. (2019).
\newblock Modeling the duration of word-final s in english with naive
  discriminative learning.
\newblock {\em {Journal of Linguistics}}, 57(1):123--161.

\bibitem[Wood, 2017]{Wood:2017}
Wood, S.~N. (2017).
\newblock {\em {Generalized Additive Models}}.
\newblock Chapman \& {H}all/{CRC}, New {Y}ork.

\bibitem[Yang, 1994]{yang1994morphosyntactic}
Yang, B.~S. (1994).
\newblock {\em Morphosyntactic phenomena of Korean in role and reference
  grammar: psych-verb constructions, inflectional verb morphemes, complex
  sentences, and relative clauses}.
\newblock PhD thesis, {State University of New York}, Buffalo.

\end{thebibliography}

\end{document}